# Military Simulator - A Case Study of Behaviour Tree and Unity based architecture

Shruti Jadon
JIIT, Noida, India

Anubhav Singhal
JIIT, Noida, India

Suma Dawn
Dept. of CSE& IT
JIIT, Noida, India

## ABSTRACT
In this paper we show how the combination of Behaviour Tree and Utility Based AI architecture can be used to design more realistic bots for Military Simulators. In this work, we have designed a mathematical model of a simulator system which in turn helps in analyzing the results and finding out the various spaces on which our favorable situation might exist, this is done geometrically. In the mathematical model, we have explained the matrix formation and its significance followed up in dynamic programming approach we explained the possible graph formation which will led improvisation of AI, latter we explained the possible geometrical structure of the matrix operations and its impact on a particular decision, we also explained the conditions under which it tend to fail along with a possible solution in future works.

## 1. INTRODUCTION
War games are the closest training to actual combat that the soldiers can go through. It recreates every physical aspect of the combat example fatigue, anxiety, smells, sound etc. But it is also very expensive. For different terrains different kinds of training camps need to be prepared and they cannot be modified easily. There are also ethical dilemmas like placing old people or children in the training. But if a soldier just needs to test his tactics or know the lay of the land, this type of training is not a necessity in initial phase. The military simulators are used for these purposes. The biggest need for the simulator is to create an enemy who reacts like a human enemy. The role of Artificial Intelligence comes into play here [1] and [10].

In game theory, to analyze any situation we define the matrix consisting of the probable reaction of the opponent to particular decision. Here in this game, we will firstly define the matrix on ideal situation which will change as time passes with help of AI designed in it [8], the iterative method of testing the correct value of the matrix will help us to achieve the more accurate values of the matrix, and this can only be achieved with more usage of this game by various communities, more specifically by defense agencies

## 2. BACKGROUND STUDY
There had been various kinds of problems in training soldiers in war games. The cost of these games is too high. A location has to been searched and built according to the specifications. This costs a lot of resources in terms of time, money, and man-power. Apart from it the ammunition and other weapons used in training add to the costs. The cultural authenticity also needs to be created especially for war in foreign countries like Iraq or Iran. Role-players generally help in achieving this goal. But then there is the ethical dilemma of keeping children and elderly as role-players. But the soldiers will have to face these situations in the battlefield. And the role-players also get fatigued and may not be able to perform the same throughout the training. This put the whole training in jeopardy. And the costs of these trainings make it difficult for everyone to get this training or train for extended period of time.

Military simulators [3], [4] and [6] have become very popular with governments within the last few years due to increase in their effectiveness in training soldiers by providing them with realistic environments. They can provide culturally authentic role-players who will never deviate from their path and they won't get fatigued or hungry. Hence the training can continue for extended periods. There will be no ethical dilemmas in virtual environment.

But taking all this into consideration, the AI of these simulators should be fully autonomous and responsive. It should be believable enough so as to make the soldiers feel that they are talking to humans

## 3. MATHEMATICAL MODEL
### 3.1 Basic Mathematical Model
The strategy matrix [11] can be stated as,

$$A.X = R \qquad (1)$$

where A is M x N matrix where N is the number of strategies used in game & M=1 (as initial AI designed [8] indicates only one strategy of the opponent) R is desired result (maximum or minimum value), As we know that the matrix column values can be represented as a vector in a space, so we can see that the vector 'X' will be some continuous curve in n-dimensional space, and the vector 'A' will be n points spread in that space, so this equation can be seen as,

$$\begin{pmatrix} a_{11} & a_{12} & \ldots & a_{1n} \end{pmatrix} \begin{pmatrix} b_{11} \\ b_{12} \\ \vdots \\ \vdots \\ b_{1n} \end{pmatrix} \qquad (2)$$

Here the values of the matrix can't be negative; its minimum value will be 0. And this can also be represented as

$$[a_1 \; a_2 \; a_3]\begin{bmatrix} b_1 \\ b_2 \\ b_3 \end{bmatrix} = a_1b_1 + a_2b_2 + a_3b_3 \quad (3)$$

So to maximize the health conditions we will put A as health matrix and B as the emphasis matrix and thus we will get numerous ways in which of such conditions which will be represented geometrically. Likewise, we can also minimize the loss condition, by replacing 'A' matrix with the loss matrix and we can find the ways to minimize them.

## 3.2 Dynamic programming in diagrammatic approach:

In this, first of all define the rankings of the strategies and then assuming the idealistic condition, draw the recursive graph of it, and thus it can be traversed randomly by the designed AI architecture.

Let the matrix A have the rankings as the places on which they are defined so we can obtain the diagram of it in Fig 1.

Likewise it can be extended and the accurate one will be determined on basis of probability and that will also result in a matrix formation (game theory).

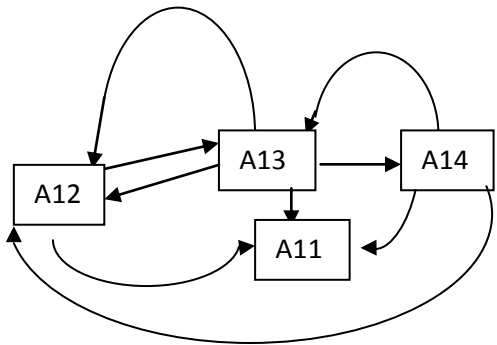

**Fig. 1 defining the ranking strategy**

Equation 1 may also be seen in terms that X is the emphasis matrix and R is desired result. So if we multiply A and X we will find a point and that can be formed by numerous combinations of the values of vector X, this will result in the favorable area in n-dimensional space of a particular situation, following this up we can find the probability of that particular situation with formula,

$$P(R) = \frac{Possible \; combinatin \; of \; X}{Total \; combination \; of \; X} \quad (4)$$

Total combinations of X also depends on vector A, the value of 'A' in turn will affect the Area. For example if the 'A' matrix consists of the m 0's[8] in it then it conclude that the numbers of points in n- dimensional space are reduced to 'n-m' number of points, and thus we can see that still we will find numerous possible combinations in X, but as to minimize the loss of money and time, we will find only 1 point in that space which consists of the 0 value of X vector which will multiply with value in A vector, and thus the curve of X will be discontinuous, in such situations it is hard to determine the actual probability. The matrix A can be defined in form of basis of a vector space and similarly the X matrix, so if let say

matrix A is, $\begin{bmatrix} 1 \\ x \\ 2x \end{bmatrix}$

And X is $[x+y \; \; x \; \; xy]$

Then their multiplication can led to, a 2 dimensional equation,

$$\begin{bmatrix} 1 \\ x \\ 2x \end{bmatrix}[x+y \; \; x \; \; xy] = x + y + x^2 + 2x^2 y.$$

As we can see the equation, it can easily get maximized or minimized if we known certain limit values of x, which is presumed to be known. But it will be quite impossible to define it, when there are numerous variables of high order in X, say

$[x+y^8, y^6z, y^3 + n^2]$ Then it will result in

$$R = x + y^8 + xy^6 z + 2xy^3 + 2xn^2.$$

It's hard to even assume how this figure might look like and what will be possible combinations of it.

It's hard to even assume how this figure might look like and what will be possible combinations of it. In this case we can use help of tropical mathematics.

Tropical geometry, is a branch of mathematics that deal with n dimensional figures, It converts complex equations in to piecewise linear in fact provide us easiness to work on enumeration of both complex and algebraic curves simultaneously. Tropical diagrams [7] can be made and analyzed in two ways first is Tropical graph made on Euclidean plane and second one is made on Tropical plane.

Through this, we can figured any curve in to a polyhedral and further we can find the possible maximum and minimum values, which gives us a rough sketch of the possible combinations of the favorable area. And thus we can make our decision, likewise.

## 4. SIMULATION ARCHITECTURE

The basic AI framework is of a behavioural tree with each level in the tree corresponding to different state the AI will be in. The state the AI will go in from transition from one level to another will be calculated by Utility Based AI [1], [3] and [4].

## 4.1 Behaviour Architecture

It is a hierarchical approach to map the behaviour of an object. On first evaluation they start from the root node and each child is evaluated. The child may be behaviour or another selector which will again evaluate which behaviour to select. If all of a child node's conditions are met, its behaviour is started. When a node starts a behaviour process, that node is set to 'running' status, and it returns the behaviour. The next time the tree is evaluated, when it comes to a 'running' node, it knows to pick up where it left off. The node can have a sequence of actions and conditions before reaching an end state. If any condition fails, the traversal returns to the parent.

BT relies on Boolean conditions for their selectors. This makes the understanding and implementation very simple but it limits the complexity of AI which can be designed through the BT. Hence we want to retain the hierarchical and simplistic approach of the BT, simultaneously being able to implement much more complex architecture.

## 4.2 Utility Based AI

There are two factors which affect the decision making: selectors and reasoners. Selectors only choose one event at a time on the basis of criteria, without calculating the probability. Whereas, Reasoners calculate the probability of many events at a time, and on the basis of their value it chooses one event.

So, Instead of selectors, it uses reasoners. The selectors use simple logic like selecting first option that fits the requirement or selecting according to predefined priority or probability.

But the reasoner is complex. It can be used to dynamically calculate weights, scores or probabilities on basis of which the choices are made[6]. There may not be a single correct choice but multiple number of choices of which one is selected based on external factors like threat, heath etc.

## 4.3 Decision making

In Fig. 1 is a very rudimentary behaviour tree of an AI in a First Person Shooter games.

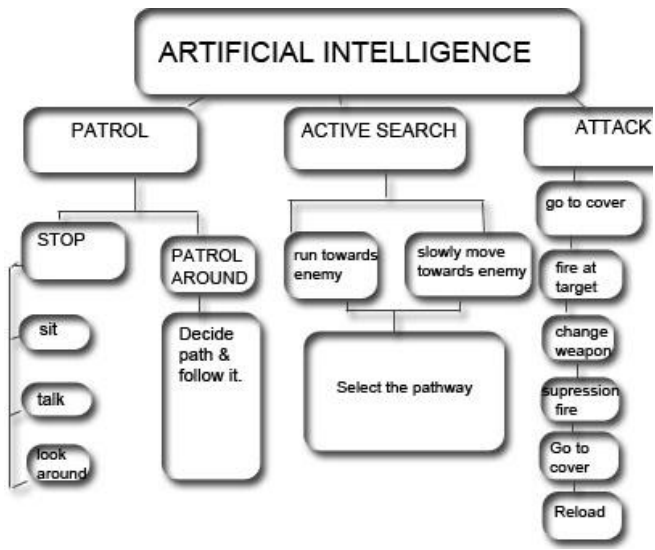

**Fig. 2 – Diagram of the Behaviour tree structure**

Initially the AI will be in patrol state. He is not actively seeking for enemy. He may go on patrol or sit around and take rest and chat with people. This subtree can easily be programmed using BT as the probability of doing any action in the subtree is equal. Therefore no complex calculations are required.

When the game AI hears a gunshot or hears a suspicious voice, he will go into the Active search state.

States are changed on the bases of a global parameter *force*. It increases when activity occurs which indicates presence of intruders like gunshots, rustling of leaves in suspicious area footsteps etc. When force reaches a threshold, the AI changes it state and goes into a more aggressive mode. Over the period of time force decreases and the AI may change back to his passive state. But when the AI is in attack mode the force value stops decreasing as enemy has been detected.

In active search it has option to use the aggressive or the stealthy option. The weights of the options are given on bases of the following equation.

$$f(x) = e^{-x} \quad (5)$$

where x= (α*Threat)+(β*health)+(γ*ammo) & x>=0 and α ,β, and γ are normalizing constants.

The weights and probabilities of the respective options will be slowly moving towards enemy: f(x)  (as shown in Fig. 2) Run towards enemy: 1-f(x)

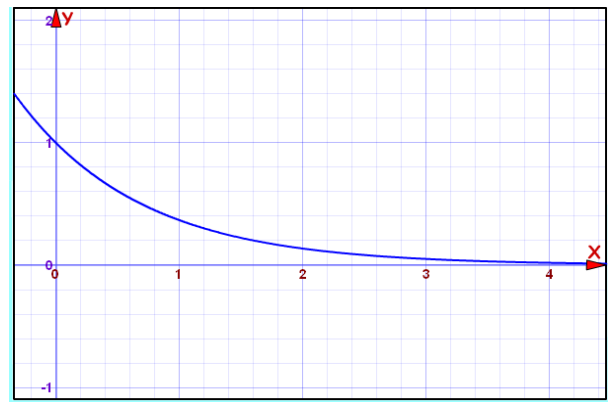

**Fig. 3 Graphical representation of probability.**

This is called marginal utility [1] and [3] because as the amount of threat increases, it becomes of much more importance than before. It is not increasing linearly. Hence the response will be much more aggressive. And if the AI has more health and ammunition, he is likely to take much more aggressive actions.

On detection of enemy AI will move into attack mode. Again it has two options. Fire from cover or to directly fire.

Again the options will be based on the equation no. (5).

Fire : $f(x)$ (as shown in Fig. 2)

Go to cover: $1 - f(x)$

As the threat increases, the probability on going into cover will also increase. The equation used above can be modified based on the requirements and the number of parameters which need to be used. If the option of going to cover is selected, then we get four options. To get their probability we can calculate their utility score. Utility score, as the name suggests tells us the necessity of a behaviour at that instance. For example, the utility score of reload can be calculated through equation

$$f(x) = e^{(1/x)} \quad (6)$$

where x denotes the amount of ammunition left in magazine. We can see as the ammo is decreasing utility is increasing. Assuming calculated utility score of all the option are $u_1, u_2, u_3, ..., u_n$ the probability(P) of each option as

$$P = u_i / (u_1 + u_2 + u_3 \cdots u_n) \qquad (7)$$

where (i=1,2,3...,n).

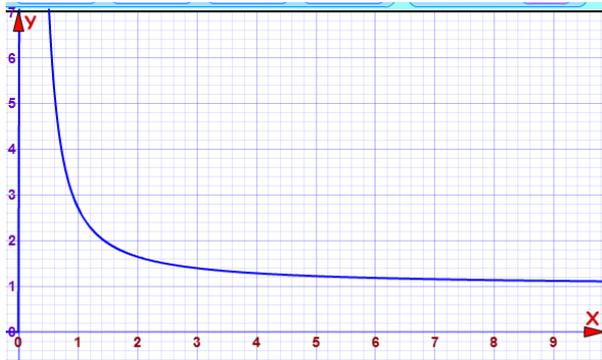

**Fig. 4 Graph for utility score of reload**

## 5. CONCLUSIONS

We have proposed an AI approach for a military simulator which is a mix of behaviour tree and utility based AI architectures. Instead of being static, the probabilities are being calculated dynamically based on the situation. Also, with the help of tropical mathematics the favorable curve design can be made and we can try to find out some specific pattern in to it, and put that data in to AI to make it more effective and optimized.